\documentclass{article}
\usepackage{amsmath,graphicx,mlspconf,amsfonts}
\makeatletter
\def\ps@camera{\let\@mkboth\@gobbletwo
  \def\@oddhead{\mbox{\footnotesize\MakeUppercase{\@toappear}}}
  \def\@oddfoot{\parbox{\textwidth}{\small\raggedright\noindent\@copyrightnotice}}}
\if@camera\ps@camera\fi
\makeatother
\usepackage{times}
\usepackage{xcolor}
\usepackage[hyperfootnotes=false]{hyperref}
\usepackage{booktabs}
\usepackage{footmisc}
\renewcommand{\footnotesize}{\fontsize{9pt}{10pt}\selectfont}

\copyrightnotice{979-8-3195-0884-3/26/\$31.00~{\copyright}2026~IEEE.\ Personal use of this material is permitted. Permission from IEEE must be obtained for all other uses, in any current or future media, including reprinting/republishing this material for advertising or promotional purposes, creating new collective works, for resale or redistribution to servers or lists, or reuse of any copyrighted component of this work in other works.}

\toappear{2026 IEEE International Workshop on Machine Learning for Signal Processing, Sep.\ 28-- Oct.\ 1, 2026, Atlanta, USA}

\title{Model-Agnostic Retrieval-Augmented Extended Forecasting for Time Series}
\name{Juan Pablo Villa Serna \quad Rohan Asthana \thanks{Rohan Asthana  is partially supported by the Celtic-Next initiative and Project SUSTAINET inNOvAte (Project ID C2024/3-4)}  \quad Vasileios Belagiannis}
\address{%
    Friedrich-Alexander-Universit\"at Erlangen--N\"urnberg \\
     Email: \{juan.pablo.villa, rohan.asthana, vasileios.belagiannis\}@fau.de
}

\begin{document}

\maketitle

\begin{abstract}
Time series forecasting with pretrained foundation models has demonstrated strong zero-shot capabilities. However, achieving optimal performance on time series with short or negligible historical data in domain-specific applications typically requires adaptation via either fine-tuning or RAG. While fine-tuning is effective, it incurs substantial computational costs. This work explores RAG within univariate time series (Retrieval Augmented Generation) as a more efficient alternative, in particular RAF \cite{b6} (Retrieval Augmented Forecasting), and introduces RAEF (Retrieval-Augmented Extended Forecasting), a model-agnostic method built upon RAF.
RAEF incorporates key refinements to the retrieval and aggregation mechanisms: (1) direct retrieval in input-space rather than embedding-space, reducing inference overhead, and (2) concatenation-based aggregation that preserves temporal structure instead of averaging. Empirical evaluation across multiple benchmark datasets demonstrates that RAEF outperforms RAF in both accuracy and inference overhead. Furthermore, comprehensive comparisons with zero-shot and fine-tuned foundation models show that RAEF achieves competitive or superior performance to fine-tuning while avoiding its computational burden, establishing it as a practical and scalable approach for domain adaptation in time series forecasting.
\end{abstract}
\begin{keywords}
RAF, RAG, Foundation Model, Time Series, Fine-tuning.
\end{keywords}

\newcommand{\cem}[1]{\textcolor{blue}{cem: #1}}
\section{Introduction}
\label{sec:intro}

\begin{figure}[h]
\centering
\includegraphics[width=0.70\linewidth]{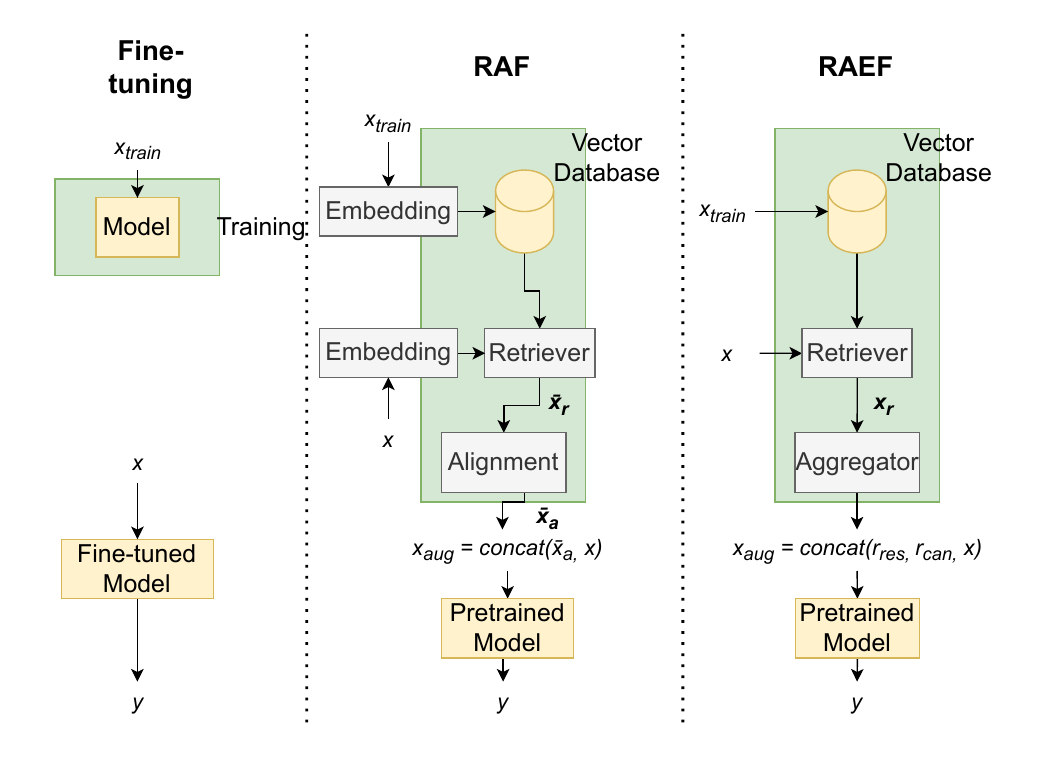}
\caption{\textbf{Architecture.} \textbf{Left:} Conventional fine-tuning, where model parameters are updated during training using $x_{\text{train}}$. \textbf{Middle:} RAF, in which $x_{\text{train}}$ is used to construct a vector database. At inference, the input $x$ is embedded to retrieve the most similar samples; their average $\bar{x}_r$ is aligned with $x$, concatenated to form $x_{\text{aug}}$, and fed into a pretrained model. \textbf{Right:} RAEF, where similarity is computed directly in input-space rather than embedding-space. Retrieved samples are aggregated to construct $x_{\text{aug}}$, effectively replacing RAF's Alignment block with an Aggregator block, which is subsequently passed to a pretrained model to predict $y$.}
\label{fig:RAEF_architecture}
\end{figure}

Time series forecasting is critical for decision-making across diverse domains, including finance, energy systems, healthcare, and climate science. Recently, transformer-based foundation models have demonstrated impressive zero-shot forecasting capabilities~\cite{b1, b2,b3}. Notable examples include Moirai~\cite{b10}, Chronos~\cite{b13}, MoiraiMoE~\cite{b20}, PatchTST~\cite{b11}, TimeGPT-1~\cite{b2}, TimesFM~\cite{b14}, GTT~\cite{b15}, Time-LLM~\cite{b16}, UniTime~\cite{b17}, UniTS~\cite{b18}, ChatTime~\cite{b21}, OFA~\cite{b22}, and Time-MoE~\cite{b24}, which leverage attention mechanisms~\cite{b9} to capture long-range temporal dependencies. These models, however, face challenges in scenarios with limited historical context. When time series data provide insufficient context windows, zero-shot approaches struggle to achieve satisfactory accuracy. In such cases, domain adaptation becomes necessary, typically accomplished through either fine-tuning or Retrieval-Augmented Generation (RAG).

Fine-tuning updates model parameters using domain-specific data, but this incurs substantial computational costs that are proportional to the model size and the scale of the dataset. In contrast, RAG, which originated in natural language processing (NLP)~\cite{b5}, offers an alternative by augmenting model inputs with relevant retrieved examples without updating the parameters. During inference, relevant samples from a prebuilt domain-specific database augment the input, thereby enabling the model to attend over extended context windows and capture domain-relevant temporal dependencies without requiring retraining.

While RAG has been extensively studied in the context of NLP, it remains relatively  less explored for time series. Recent methods including TimeRAF (Retrieval Augmented for Time Series Forecasting)~\cite{b7}, RATD (Retrieval-Augmented Diffusion Models for Time Series Forecasting)~\cite{b8}, RAFT (Retrieval Augmented Time Series Forecasting)~\cite{b30}, and TS-RAG (Retrieval-Augmented Generation based Time Series Foundation)~\cite{b33} require custom architectures or domain-specific training, which limits their model-agnostic applicability. Only RAF (Retrieval Augmented Forecasting)~\cite{b6} and FinSrag (Financial time-Series RAG)~\cite{b32} maintain model-agnostic properties. FinSrag relies on large language models, which results in a computational overhead that is comparable to that of fine-tuning. RAF retrieves similar time series samples using embeddings and averages their future components before concatenation with the query input. Nevertheless, this averaging strategy can dilute important temporal patterns, thus limiting its effectiveness.

We present Retrieval-Augmented Extended Forecasting (RAEF), a model-agnostic approach for time series forecasting that addresses the limitations of RAF with two key refinements. Firstly, we operate on the input signal rather than embedding-space, as in RAF. This modification reduces retrieval time by a factor of four across context lengths of $32, 64, 128$. Secondly, we preserve the temporal structure of the data by concatenating the retrieved samples rather than averaging them. To evaluate RAEF, we conduct a comprehensive evaluation across six benchmark datasets using three foundation models (MoiraiMoE, Chronos-T5, and Chronos-Bolt). We are the first to offer a comparison between RAG and model fine-tuning in the context of time series. We demonstrate an improvement in MASE \cite{b34} of 11--16\% over the base model. RAF achieves a more modest improvement of 1--7\% over the same base model, whereas RAEF delivers results that are comparable to or better than those of fine-tuning. In summary: 1) RAEF operates in input space, 2) RAEF discards RAF's alignment step in favor of an aggregator module, and 3) we evaluate against fine-tuning, and are the first RAG method for time series to surpass it.

\section{Method}
\label{sec:II}

Given a univariate time series sample $x \in \mathbb{R}^{C}$ with a length $C$, which we refer to as the context, we aim to predict the future sequence $y \in \mathbb{R}^{F}$ of length $F$, referred to as the future. To achieve this, we construct a vector database from training samples $x_{\text{train}} \in \mathbb{R}^{C+F}$ and leverage these samples containing domain-specific information during inference to augment the input $x$, yielding the augmented input $x_{\text{aug}}$. Subsequently, $x_{\text{aug}}$ is fed through a pretrained foundation model to produce the prediction $y$. Our method builds upon RAF, introducing two key modifications detailed in the following subsections.

\subsection{RAF}
\label{subsection:RAF}
RAF is illustrated in the middle panel of Figure~\ref{fig:RAEF_architecture}. Instead of further training the foundation model, samples $x_{train}$ are split between context $d$ and futures $p$, the context is embedded and ingested into a vector database $D$. At inference time, the query sample $x$ is embedded, and retrieval from the database (retriever) is performed by computing the Euclidean distance between the query embedding and stored embeddings.

The retriever identifies the top $k$ most similar sequences based on Euclidean distance in feature space and retrieves contexts $r \in \mathbb{R}^{C \times k}$ with their respective futures $f \in \mathbb{R}^{F \times k}$ and scores $s \in \mathbb{R}^{1 \times k}$. The context and future pairs are concatenated along the temporal dimension to form $x_r$, and the $k$ retrieved samples are aggregated by averaging to obtain a single representative sequence $\tilde{x}_r \in \mathbb{R}^{(C + F) \times 1}$. Subsequently, instance normalization~\cite{b12} with respect to $x$ is applied to both $\tilde{x}_r$ and $x$ itself, setting the mean to zero and the standard deviation to one to mitigate distribution shift. RAF then performs alignment between the normalized $\tilde{x}_r$ and $x$ prior to concatenation. Alignment removes discontinuities at the boundary between  $\tilde{x}_r$ and $x$ by shifting one sequence along the vertical axis such that the last point of $\tilde{x}_r$ matches the first point of $x$. After normalization, alignment and concatenation, $x_{\text{aug}}$ is formed and subsequently fed into the pretrained foundation model to produce the output $y$. To recover the output in the original input scale, $y$ is denormalized to obtain the final forecast.

While this approach is effective, the averaging operation on retrieved samples results in information loss, thereby limiting the model’s ability to fully exploit the information from all retrieved examples. Furthermore, the application of embeddings at each inference step introduces computational overhead that degrades overall performance. Time series are already real-valued sequences living in a metric space where Euclidean distance directly captures similarity, making additional projection unnecessary and potentially lossy.

\subsection{Retriever}

Our approach employs the same retrieval mechanism as RAF, with the key distinction that the embedding step is omitted. Time series inputs are already continuous numerical vectors where direct distance metrics are well-defined. Additional embedding layers introduce computational overhead without guaranteed accuracy improvements and may even degrade performance if poorly suited to the domain. By operating directly in input space, RAEF reduces inference overhead while preserving retrieval quality.

\subsection{Aggregator}
\label{subsection:Aggregator}

The same instance normalization as RAF is applied on $x$, obtaining $\tilde{x}$, mean $\mu_{\mathbf{x}}$ and standard deviation $\sigma_{\mathbf{x}}$. The normalized retrieved samples $\tilde{x_r}$  and scores $\tilde{s}$ are calculated as:

\begin{equation}
\tilde{x_r} = \frac{x_r - \mu_{\mathbf{x}}}{\sigma_{\mathbf{x}}}, \tilde{s} = \frac{s}{\sigma_{\mathbf{x}}^2 \cdot C},
\label{equation:score_norm}
\end{equation}

 where $s$ corresponds to a sum of squared differences over context length $C$ and $x_r$ is calculated as obtained in RAF. We normalize the scores $s$ by dividing with ${\sigma_{\mathbf{x}}}^2 \cdot C$ to obtain a scale-invariant distance $\tilde{s}$.

Our objective is to aggregate all retrieved samples rather than averaging them, thereby preserving the complete retrieved information. However, this approach generates long sequences at each inference stage, increasing the computational burden on the pretrained model when attending to these extended sequences. To find a balance between information retention and computational efficiency, we define a score threshold $d_t$ that discriminates between retrieved candidates $r_{can}$ (high relevance) and retrieved residuals $r_{res}$ (low relevance).

\begin{equation}
\begin{aligned}
\mathcal{C}
&=
\left\{
\tilde{x}_r, \;\middle|\;
\tilde{s} \le d_t
\right\}
\\
\mathcal{R}
&=
\left\{
\tilde{x}_r, \;\middle|\;
\tilde{s} > d_t
\right\}
\end{aligned}
\label{equation:discriminator}
\end{equation}

\begin{equation}
\begin{aligned}
r_{\mathrm{can}}
&=
\operatorname{concat}_{x \in \mathcal{C}}(x)
\end{aligned}
\label{equation:concatenation}
\end{equation}

\begin{equation}
\begin{aligned}
\hspace{-1.3em}
r_{\mathrm{res}}
&= \frac{1}{|\mathcal{R}|}
\sum\limits_{x \in \mathcal{R}} x
\end{aligned}
\label{equation:residual}
\end{equation}

Retrieved samples are divided into candidate $\mathcal{C}$ and residual sets $\mathcal{R}$ as defined in equation~\ref{equation:discriminator}. Candidates are concatenated to form $r_{can}$, while residuals are averaged to obtain $r_{res}$. The vector $r_{can}$ contains in a single sequence the most similar samples to the input of interest, enriching the sample with the most relevant patterns. By contrast, $r_{res}$, despite having lower relevance, is not entirely discarded but averaged. At the end $d_t$ helps to balance between very long augmented sample with less relevant segments and very short augmented sample with lost information due to averaging across many samples. We did not discard the residual sets because these samples still contain useful information, as this is how RAF performs augmentation. The final augmented input $x_{aug}$ is constructed as:

 \begin{equation}
\begin{aligned}
x_{\mathrm{aug}}
&=
\operatorname{concat}
\left(
r_{\mathrm{res}},
r_{\mathrm{can}},
\tilde{x}
\right).
\end{aligned}
\label{equation:aggregator}
\end{equation}
The retrieved samples are concatenated in ascending score order to preserve similarity-based relevance. When either the residual set or the candidate set is empty, $x_{aug}$ is constructed using only the available set.

Additionally, as part of our improvements, we removed the alignment step introduced in RAF. Originally, the alignment was purposefully added to eliminate discontinuities between the retrieved samples and the input of interest. However, we propose that by removing this alignment, the retrieved samples will be treated by the pretrained model as past periodic representations. Given that the retrieved samples $r_{\text{can}}$ exhibit high similarity with $x$, the augmented sample $x_{\text{aug}}$ effectively represents a longer semi-periodic sequence, allowing the model to capture extended temporal patterns more naturally.

\section{Experiments}
\label{sec:III}

Our experiments aim to demonstrate that RAEF: (1) improves forecasting accuracy over baseline foundation models and RAF, (2) achieves better performance than fine-tuning without parameter updates, and (3) benefits from input-space retrieval and structure-preserving aggregation.

\subsection{Experimental Setup}
We evaluate our method on six benchmark datasets spanning diverse domains {\href{https://github.com/zhouhaoyi/ETDataset}{[ET]}, \href{https://www.kaggle.com/datasets/dharanikra/electrical-power-demand-in-turkey}{[Power]}, \href{https://www.kaggle.com/datasets/leonardo00/istanbul-traffic-index}{[Traffic]}, \href{https://zenodo.org/records/4654833/files/fred_md_dataset.zip}{[FredMd]}, \href{https://archive.ics.uci.edu/static/public/321/electricityloaddiagrams20112014.zip}{[ElectricityUCI]}, \href{https://github.com/sir-lab/time-series-fm-dataset.git}{[Huawei Cloud]}} using split sizes of $0.8$ and instance normalization as preprocessing step. MoiraiMoE~\cite{b20} serves as our primary backbone (inference time: 16ms), with additional evaluation on Chronos-T5~\cite{b13} and Chronos-Bolt to demonstrate model-agnostic applicability. Our benchmark includes main zero-shot foundation models, we excluded common models like PatchTST~\cite{b11} due to lacking zero-shot capabilities. Our focus is on enhancing the performance of zero-shot foundation models through RAG.

We use MASE (Mean Absolute Scaled Error) utilized in RAF as our evaluation protocol. MASE normalizes prediction error by naive forecast error, enabling fair comparison across datasets. We fix forecasting horizon at $F=16$ (MoiraiMoE's maximum non-autoregressive length) to avoid accumulative error \cite{b28} and evaluate context lengths $C \in \{32, 64, 128\}$. Vector databases contain $10,000$ training samples, constructed using HNSW~~\cite{b27} within the Chroma framework \footnote{\url{https://www.trychroma.com/}}, and take approximately 5 minutes to build. Empirical results demonstrate that a threshold of $d_t=1.0$  is effective across our experimental settings (see table~\ref{tab:s_mid}). All experiments use three random seeds with identical train/test splits. Within the MASE results, the standard deviation is within $0.001$ to $0.01$. We set $k=16$ as it provides sufficient retrieved context without exceeding the model's effective attention window.

We compare against: (1) \textbf{Base}: Pretrained model without adaptation, (2) \textbf{Fine-Tuning (FT)}: Full parameter updates with early stopping, (3) \textbf{RAF}~\cite{b6}: embedding-space retrieval with alignment and averaging, (4) \textbf{RAEF}: Our method. Fine-tuning uses learning rate of $10^{-4}$, batch size of $1024$, weight decay of $10^{-2}$, with betas of $0.9,0.98$. We implemented early stopping based on validation accuracy with a patience of $3$, up to maximum of $500$ iterations, same optimizer as MoiraiMoE and same $x_{train}$ as training set. RAEF and RAF require no parameter updates. Experiments are reported in \footnote{\url{https://github.com/jpvilla1990/raef/}}

\subsection{Results}

Table~\ref{tab:results} presents our main results across all datasets, foundation models, and context lengths. RAEF consistently achieves the best performance among MoiraiMoE variants, with average improvements of $15.80\%$, $11.78\%$, and $11.26\%$ for contexts $32$, $64$, and $128$, respectively, over the baseline. Notably, RAEF outperforms fine-tuning in most cases while requiring no parameter updates, likely because transformer-based models are inherently trained to attend over long sequences.

\begin{table}[htbp]
\centering
\scriptsize
\caption{MASE for context sizes 32, 64, and 128 across different foundation models. FT = Fine-Tuning. Bold indicates best method on MoiraiMoE, underlining indicates absolute best. On Chronos T5 and Bolt, × indicates datasets used during model pretraining and therefore excluded from evaluation.}
\label{tab:results}
\vspace{2mm}
\setlength{\tabcolsep}{3pt}
\resizebox{\columnwidth}{!}{
\begin{tabular}{l|cccc|ccc|ccc}
\toprule
 & \multicolumn{4}{c|}{MoiraiMoE} & \multicolumn{3}{c|}{Chronos Bolt} & \multicolumn{3}{c}{Chronos T5} \\
Dataset & Base & FT & RAF & RAEF & Base & RAF & RAEF & Base & RAF & RAEF \\
\midrule
\multicolumn{11}{l}{\textbf{Context 32}} \\
\midrule
ET             & 1.0414 & 0.9990 & 0.9672 & \textbf{\underline{0.8944}} & 0.9284 & 0.9566 & 0.9021 & 0.9816 & 0.9601 & 0.8999 \\
Huawei Cloud   & 1.2000 & 1.1355 & 1.0975 & \textbf{0.9872} & 1.3591 & 1.2809 & 1.1076 & 1.0970 & 1.0244 & \underline{0.8585} \\
power          & 1.1936 & 1.1833 & 1.0251 & \textbf{0.9324} & × & × & × & × & × & × \\
traffic        & 0.1734 & 0.1732 & \textbf{0.1655} & 0.1726 & × & × & × & × & × & × \\
fredMd         & 0.8565 & 0.8410 & 0.8341 & \textbf{0.7381} & × & × & × & × & × & × \\
electricityUCI & 2.9924 & 2.8827 & 2.9066 & \textbf{2.6647} & × & × & × & × & × & × \\
\midrule
Improvement (\%) & 0.00 & 2.75 & 7.31 & \textbf{15.80} & 0.12 & 0.64 & 5.95 & 3.87 & 6.40 & \underline{13.88} \\
\midrule
\multicolumn{11}{l}{\textbf{Context 64}} \\
\midrule
ET             & 0.9389 & 0.9111 & 0.9452 & \textbf{0.8590} & 0.8797 & 0.9124 & \underline{0.8555} & 0.9021 & 0.9207 & 0.8685 \\
Huawei Cloud   & 1.0168 & 1.0313 & 0.9907 & \textbf{0.8876} & 1.2519 & 1.2256 & 1.2058 & 0.8871 & 0.9259 & \underline{0.8153} \\
power          & 0.9729 & \textbf{0.9702} & 1.0002 & 0.9814 & × & × & × & × & × & × \\
traffic        & 0.1578 & 0.1587 & \textbf{0.1547} & 0.1625 & × & × & × & × & × & × \\
fredMd         & 0.7451 & 0.7040 & 0.7143 & \textbf{0.6112} & × & × & × & × & × & × \\
electricityUCI & 2.6482 & 2.3876 & 2.5151 & \textbf{2.0586} & × & × & × & × & × & × \\
\midrule
Improvement (\%) & 0.00 & 3.02 & 1.81 & \textbf{11.78} & -3.01 & -3.53 & -1.48 & 4.68 & 2.95 & \textbf{8.21} \\
\midrule
\multicolumn{11}{l}{\textbf{Context 128}} \\
\midrule
ET             & 0.8689 & 0.8548 & 0.8623 & \textbf{0.7962} & \underline{0.7561} & 0.7983 & 0.7645 & 0.7798 & 0.7881 & 0.7609 \\
Huawei Cloud   & 0.9986 & 0.9434 & 0.9660 & \textbf{0.8632} & 1.2630 & 1.1472 & 0.9955 & 0.9371 & 0.9004 & \underline{0.7197} \\
power          & 0.9210 & 0.9179 & \textbf{0.9050} & 0.9099 & × & × & × & × & × & × \\
traffic        & 0.1576 & 0.1581 & \textbf{0.1571} & 0.1595 & × & × & × & × & × & × \\
fredMd         & 0.9076 & 0.8829 & 0.8965 & \textbf{0.7749} & × & × & × & × & × & × \\
electricityUCI & 2.4884 & 2.4167 & 2.5585 & \textbf{1.9808} & × & × & × & × & × & × \\
\midrule
Improvement (\%) & 0.00 & 2.20 & 0.78 & \textbf{11.26} & -1.50 & -1.03 & 3.50 & 4.50 & 5.29 & \textbf{13.24} \\
\bottomrule
\end{tabular}
}
\end{table}

\subsubsection{Model-Agnostic Performance}
Results across Chronos-T5 and Chronos-Bolt demonstrate RAEF's model-agnostic properties. For Chronos-T5, RAEF achieves $13.88\%$, $8.21\%$, and $13.24\%$ improvements across context lengths, with particularly strong performance on Huawei Cloud. For Chronos-Bolt, which exhibits strong baseline performance, RAEF still achieves $3.50\%$ improvement at context $128$, confirming its general applicability. 

\subsubsection{Context Length Analysis}
As hypothesized, improvement decreases with longer contexts. Shorter contexts ($C=32$) benefit most from retrieval augmentation ($15.80\%$ for MoiraiMoE), since limited temporal information leaves more room for retrieved samples to contribute additional patterns. Longer contexts  ($C=128$) already encode sufficient information, reducing marginal gains from augmentation to $11.26\%$. This trend holds across all foundation models, making RAEF particularly valuable in data-scarce scenarios where historical context is limited. Extending further, average improvement dropped to $6\%$ at $C=256$ and $-0.07\%$ at $C=512$. We do not claim an absolute threshold where augmentation stops helping, but the tendency is clear: shorter contexts benefit most from RAG on time series.

\subsubsection{Comparison with Fine-Tuning}
RAEF achieves comparable or superior performance to fine-tuning on $4$ out of $6$ datasets while requiring no parameter updates, making it more practical for rapid domain adaptation. Fine-tuning's advantages on Power and Traffic datasets are marginal ($0$-$1\%$ improvement over RAEF), suggesting these datasets' patterns are already well-captured by pretrained models. The combination of competitive accuracy with zero training overhead positions RAEF as an efficient alternative to fine-tuning in resource-constrained environments.

\subsubsection{Aggregation Mechanism}
The performance gains can be attributed to RAEF's concatenation based aggregation, which preserves individual temporal sequences rather than averaging them as in RAF. Averaging dilutes temporal patterns across sequences, causing information loss that limits the model's ability to extract relevant features. In contrast, concatenation allows the attention mechanism to selectively attend to relevant patterns from each retrieved sample. The threshold-based candidate-residual separation further enhances this by prioritizing high-similarity samples ($d_t \leq 1.0$) for concatenation while averaging only distant, less-relevant samples, leveraging the pretrained model’s attention blocks to reduce computational overhead.

\subsubsection{Computational Efficiency}
Input-space retrieval scoped to $C=32, 64, 128$ reduces overhead from $8$ms to $2$ms per query. We observed no retrieval-time variation across context lengths.

\subsection{Ablation Studies}

\subsubsection{Aggregation Mechanism.} From Table~\ref{tab:results_components}, removing the aggregator (Keeping the alignment from RAF) still improves over standard RAF ($10.14\%$ vs $7.31\%$ at context $32$) but underperforms full RAEF. The threshold-based candidate-residual separation contributes an additional $5.66\%$ improvement on average, demonstrating the value of prioritizing high-similarity samples while not discarding lower-similarity information entirely through averaging.

\subsubsection{Hyperparameter Analysis}
We evaluate empirically the hyperparameter $d_t$ as shown in table~\ref{tab:s_mid}. Performance reaches a peak at a normalized distance $d_t=1.0$ with respect to equation \ref{equation:score_norm}, corresponding to one standard deviation from the query sample. This aligns with statistical intuition: samples within one standard deviation share strong distributional similarity. Beyond this threshold, retrieved samples no longer contribute meaningful individual information and can be safely averaged.

\begin{table}[t]
\centering
\caption{Impact of threshold $d_t$ on MASE improvement (\%) over baseline using MoiraiMoE-small with RAEF.}
\label{tab:dt_ablation}
\resizebox{0.45\textwidth}{!}{
\begin{tabular}{lcccccc}
\toprule
$d_t$ & 0.1 & 0.5 & 1.0 & 1.5 & 2.0 & 3.0 \\
\midrule
Improvement (\%) & 5.50 & 6.41 & \textbf{6.77} & 6.71 & 6.69 & 6.68 \\
\bottomrule
\end{tabular}
}
\label{tab:s_mid}
\end{table}

\begin{table}[t]
\centering
\resizebox{0.30\textwidth}{!}{
\begin{tabular}{l|ccc}
\toprule
\multicolumn{4}{c}{\textbf{Model Component Analysis.}} \\
\midrule
 & \multicolumn{3}{c}{Improvement (\%)} \\
\cmidrule(lr){2-4}
Model Configuration & Context 32 & Context 64 & Context 128 \\
\midrule
Base & 0.00 & 0.00 & 0.00 \\
+ RAF & 7.31 & 1.81 & 0.78 \\
+ RAEF + emb L2 & 12.58 & 2.96 & 3.96 \\
+ RAEF + emb cos & 3.94 & 2.79 & 3.29 \\
+ RAEF (w/o agg) & 10.14 & 5.22 & 3.94 \\
+ RAEF & \textbf{15.80} & \textbf{11.78} & \textbf{11.26} \\

\end{tabular}
}
\caption{MASE improvement (\%). emb cos = embedding-space with cosine distance, emb L2 = embedding-space with L2 distance, RAEF (w/o agg) denotes the variant without the aggregator module, RAEF denotes our method. MoiraiMoE foundation model, horizon 16.}
\label{tab:results_components}
\end{table}

Finally Table \ref{tab:results_components} shows how the aggregator contributes to the improvement in accuracy.

\section{Conclusions}
\label{sec:V}
We introduced RAEF, a model-agnostic retrieval-augmented framework that achieves significant MASE improvements ($11-16\%$) over baseline foundation models without parameter updates. Results across six datasets and three foundation models demonstrate performance comparable to or exceeding fine-tuning on 4/6 datasets, making RAEF a practical alternative for domain adaptation in resource-constrained environments.


\bibliographystyle{IEEEbib}
\bibliography{strings,refs}

\end{document}